# CXR-LLAVA: a multimodal large language model for interpreting chest X-ray images

Seowoo Lee M.D.[1], Jiwon Youn[2], Hyungjin Kim M.D.[1],

Mansu Kim Ph.D.[2†], Soon Ho Yoon M.D. Ph.D.[1†]

[1]Department of Radiology, Seoul National University College of Medicine, Seoul National University Hospital, Seoul, Republic of Korea.

[2] AI Graduate School, Gwangju Institute of Science and Technology, Gwangju, Republic of Korea.

†Co–corresponding authors.

**Co-correspondences:**

**Soon Ho Yoon, M.D.**

Department of Radiology, Seoul National University College of Medicine, Seoul National University Hospital, Republic of Korea.

Email: yshoka@snu.ac.kr

**Mansu Kim, Ph.D.**

AI Graduate School, Gwangju Institute of Science and Technology (GIST), Gwangju, Republic of Korea.

Email: mansu.kim@gist.ac.kr

**Type of manuscript:** Original Research

**Word count for text:** 3252

# CXR-LLAVA: a multimodal large language model for interpreting chest X-ray images

Manuscript type: Original Research

# Abstract

**Purpose:** This study aimed to develop an open-source multimodal large language model (CXR-LLAVA) for interpreting chest X-ray images (CXRs), leveraging recent advances in large language models (LLMs) to potentially replicate the image interpretation skills of human radiologists

**Materials and Methods:** For training, we collected 592,580 publicly available CXRs, of which 374,881 had labels for certain radiographic abnormalities (Dataset 1) and 217,699 provided free-text radiology reports (Dataset 2). After pre-training a vision transformer with Dataset 1, we integrated it with an LLM influenced by the LLAVA network. Then, the model was fine-tuned, primarily using Dataset 2. The model's diagnostic performance for major pathological findings was evaluated, along with the acceptability of radiologic reports by human radiologists, to gauge its potential for autonomous reporting.

**Results:** The model demonstrated impressive performance in test sets, achieving an average F1 score of 0.81 for six major pathological findings in the MIMIC internal test set and 0.62 for seven major pathological findings in the external test set. The model's F1 scores surpassed those of GPT-4-vision and Gemini-Pro-Vision in both test sets. In human radiologist evaluations of the external test set, the model achieved a 72.7% success rate in autonomous reporting, slightly below the 84.0% rate of ground truth reports.

**Conclusion:** This study highlights the significant potential of multimodal LLMs for CXR interpretation, while also acknowledging the performance limitations. Despite these challenges, we believe that making our model open-source will catalyze further research, expanding its effectiveness and applicability in various clinical contexts. CXR-LLAVA is available at

https://github.com/ECOFRI/CXR_LLAVA.

# Introduction

Advances in deep learning, marked by the emergence of convolutional neural networks (CNNs) and vision transformers (ViTs), have profoundly impacted radiology [1-3]. Numerous deep learning algorithms have made their way into practical, commercial applications. However, while CNNs and ViTs are adept at specific tasks, such as classification and segmentation, this specialization could limit their ability to address multifaceted challenges in areas such as radiology. Concurrently, the natural language processing domain has witnessed significant breakthroughs, enabling large language models (LLMs), such as ChatGPT, to understand and generate human-like text with remarkable proficiency and unprecedented performance levels in linguistic tasks ranging from text generation to translation [4]. The integration of natural language processing and image processing technologies has led to the development of models that have set new benchmarks in the field, such as contrastive language-image pre-training (CLIP) [5] and the bootstrapping language-image pre-training (BLIP-2) model, which was introduced in 2023 and can interpret the context within images and generate detailed captions [6].

Most LLMs have primarily focused on text processing. However, there is a growing trend towards a multimodal approach involving processing of image, text, and even video data. OpenAI and Google have released general-purpose multimodal models (GPT-4-vision and Gemini-Pro-Vision, respectively). Furthermore, the Large Language and Vision Assistant (LLAVA), an open-source project combining vision encoding with an LLM, has demonstrated exemplary performance across a range of visual tasks [7]. However, it remains unclear how effective these general-purpose models are at interpreting chest X-rays (CXRs). Within the medical domain, there are few specific multimodal models. Google has published results for ELIXR, a model capable of interpreting CXRs, but this model is not publicly available [8]. Similarly, the open-source LLAVA-MED, a model fine-tuned for the medical domain, has been released. However, detailed insights into its proficiency in interpreting CXRs remain limited [9].

Radiologists' workload has significantly increased over the past three decades, potentially impacting the accuracy of radiologic diagnoses [10]. In response, numerous studies have explored the use of deep learning models to improve diagnostic accuracy and reduce the burden on radiologists. [11] Building on this line of research, our study employed the latest technology, a multimodal LLM, to attempt radiologic report generation for CXRs. Specifically, this study aimed to develop a multimodal LLM specifically designed for CXR interpretation, while also exploring its potential for autonomous CXR reporting.

## Materials and Methods

This study solely used publicly available datasets and did not require institutional review board approval.

**Data collection**

For model training, we included several public CXR datasets, collecting a total of 592,580 frontal CXRs (Table 1). The Medical Information Mart for Intensive Care (MIMIC) dataset provides radiologic reports in a free-text form (Dataset 2, n=217,699), while the other training datasets have multi-class or binary labeling for radiographic abnormalities (Dataset 1, n=374,881). Some datasets contain information regarding lesions' location, but this information was not utilized.

**Adapting a multimodal LLM to CXRs (CXR-LLAVA)**

A model influenced by the LLAVA network was developed (7). LLAVA, an extension of the traditional LLM, incorporates an image encoder that converts images into a sequence of image tokens. These tokens are then combined with query text tokens for text generation within the LLM. Our primary objective was to fine-tune LLAVA using CXR–radiologic report pairs. To achieve optimal performance, we developed a custom image encoder from scratch rather than using pre-trained weights. Specifically, we empirically employed the "ViT-L/16" version of the vision transformer as the image encoder. This encoder compresses information from CXRs and generates an image token sequence, which is subsequently fed into the LLM. In alignment with LLAVA's framework, we also utilized the Large Language Model Meta AI (LLAMA)-2 as our language model. [12] We selected the version with 7 billion parameters due to cost considerations.

The final CXR-LLAVA takes a CXR image and question prompt as input; the image is transformed into an image token via an image encoder, and the prompt is converted to text tokens through a tokenizer. Both are then fed into a causal language model, which autoregressively generates text responses to the questions. The trained model is available as open-source (https://github.com/ECOFRI/CXR_LLAVA), and its demo can be found at https://radiologist.app/cxr-LLAVA.

**Training step 1: constructing and training a CXR-specific image encoder**

Despite the capabilities of pretrained image encoders in understanding common visual objects, they often fall short in accurately describing radiographic findings. In this section, we propose an image encoder, $f(x)$, based on ViT-L/16 and a two-step strategy for training them to learn the radiological context specific to CXR images.

In first step, a simple classification task was used to train the image encoder (Figure 1A). The image encoder transformed a CXR image ($x_{img}$) into a representation ($z_{img}$) and then classified an abnormality ($y$) by adding a simply fully connected layer as a classifier (i.e., $y = g(z_{img})$). This classification task enabled the model to learn a fundamental yet crucial ability regarding abnormalities. We used 374,881 image-label pairs from Dataset 1 to train and validate our image encoder. We assigned binary labels: when images had labels associated with pathology, they were labeled as "abnormal," while those marked as "no finding" were designated "normal." The detailed implementation and settings are described in the supplementary material.

In the second step, the image encoder, $f(x_{img})$, was further trained based on the CLIP strategy to learn complex representations of radiological term. (Figure 1B) [5] Specifically, the image encoder, $f(x_{img})$ and pretrained text encoder, $h(x_{text}^k)$ from prior research [13] were employed to compute the contrastive loss between the image vector ($z_{img}$) and text vectors ($z_{text}^k$) (i.e., $z_{img} = f(x_{img})$ and $z_{text} = h(x_{text}^k)$). In our study, we choose $k$ keywords as $x_{text}^k$, such as "atelectasis," "pneumonia," and so forth. Using the CLIP method, the image encoder learned shared representations between images and text by mapping corresponding image and text pairs closer, and non-corresponding ones further apart. The 592,580 image-text pairs from Datasets 1 and 2 were used in the training and validating process; the detailed process is described in the supplementary material.

After completing the complex training process, our image encoder was able to understand and interpret medical images by bridging the gap between visual data and textual descriptions, potentially advancing diagnostic accuracy and efficiency.

**Training step 2: feature alignment and end-to-end fine-tuning of CXR-LLAVA**

Before fine-tuning the CXR-LLAVA model, the features from the image encoder, as described in step 1, and language model (i.e., LLaMa-2) were aligned through additional training, where the image encoder and language model weights were frozen, updating only the projection matrix ($P$). The aligned image representation (i.e., <image> = $P \times z_{img}$) was computed by updating $P$ using CXR images with refined radiologic reports (Figure 1C).

After aligning the image features, CXR-LLAVA underwent an instruction-tuning process, which was critical for refining the model's interpretative capabilities (Figure 1D). This process involved using refined radiology reports and multi-turn question-answer dialogues generated by GPT-4. These dialogues expanded upon hypothetical question-answer pairs, focusing on details such as the specific locations of anomalies, differential diagnoses, and recommendations for further radiological studies. The instruction-tuning, distinct from the initial training, focused on improving the model's ability to engage in more complex and informative interactions regarding CXRs, beyond simple image interpretation. The fine-tuning encompassed all trainable weights of the model, excluding the image encoder, aiming for a comprehensive enhancement of the model's interpretive and interactive capabilities. More details about this instruction-tuning process can be found in the supplementary materials.

**Internal and external test set composition**

For internal model testing, we utilized a randomly selected MIMIC dataset, comprising 3000 images and accompanying free text radiologic reports. [14] These were not used during the model's training and validation phases. Additionally, we employed the CheXpert test dataset, which consists of 518 images, each binary labeled for 14 findings: atelectasis, cardiomegaly, consolidation, edema, enlarged cardiomediastinum, fracture, lung lesion, lung opacity, no finding, pleural effusion, pleural other, pneumonia, pneumothorax, and support devices [15]. For external model testing, we used a dataset from Indiana University, consisting of 3689 pairs of images and free-text radiologic reports [16].

**Comparison with other multimodal LLMs**

To evaluate the performance of our model, we compared its results with those of other publicly available multimodal LLMs, including OpenAI's GPT-4-vision and Google's Gemini-Pro-Vision. Despite being in a preview state and not being specifically fine-tuned for CXR report generation, these general-purpose models have shown some potential. For instance, GPT-4-vision has demonstrated a limited ability to detect abnormalities in CXRs and the capacity to solve United States Medical Licensing Examination tests [17,18]. However, LLAVA-MED, a model fine-tuned for medical image analysis, failed to generate accurate radiologic reports from CXRs, producing nearly identical reports for diverse CXRs and was therefore excluded from our study. Other models, such as ELIXR and Med-PALM, which claim the ability to interpret CXRs, were not publicly available and thus were not included in this analysis [8,19].

During the inference process using GPT-4-vision and Gemini-Pro-Vision, we utilized their official application programming interfaces to ensure reliable outcomes. For GPT-4-vision, we used the high-resolution mode and set the model temperature to 0 to increase reproducibility. However, GPT-4-vision often rejected requests to evaluate radiologic images, suggesting that they should be assessed by a healthcare professional, which required us to modify the prompt accordingly. For Gemini-Pro-Vision, we also maintained the model temperature at 0 for consistent reproducibility. The specific prompts utilized during the evaluation process are detailed in the supplementary materials.

**Internal test set evaluation**

To evaluate the performance of radiologic report generation in the MIMIC internal test set, we utilized CheXpert-Labeler to generate pathological labels [15]. This tool analyzes free-text radiologic reports and generates labels such as positive, negative, or uncertain for each pathological finding (atelectasis, cardiomegaly, consolidation, edema, enlarged cardiomediastinum, fracture, lung lesion, lung opacity, no finding, pleural effusion, pleural other, pneumonia, pneumothorax, and support devices). We compared these labels from the model-generated reports with those from the original ground truth reports (Figure 2A).

For the CheXpert test set, which does not contain ground-truth radiologic reports, we instructed the model to generate binary labels for the same 14 findings. These labels were then compared with the ground truth. This dataset is identical to that used in a previous study where the CheXzero model exhibited expert-level pathology detection capabilities [20]. Therefore, we evaluated our model's performance against both CheXzero and the average diagnostic performance of three board-certified radiologists, as documented in the same publication (Figure 2B).

**External test set evaluation and human radiologist evaluation**

To evaluate the model's performance on the Indiana external test set, we employed the same methodology used for the MIMIC internal test set, which involved comparing the labels generated from the model's reports with the ground truth (Figure 2A).

To assess the model's capability for autonomous or semi-autonomous reporting without human radiologist intervention, an evaluation was conducted involving three human radiologists. They were presented with a set of 50 randomly selected radiographs from the Indiana external test set, comprising 25 abnormal and 25 normal images. Each radiologist reviewed 100 paired images and reports. The model-generated reports and ground truth reports were presented in a random order. They rated the acceptability of each report on a 4-point scale: A) totally acceptable without any revision, B) acceptable with minor revision, C) acceptable with major revision, and D) unacceptable. This rating system was centered around referable abnormalities, defined as findings that necessitate further examination, consultation, or follow-up (e.g., lung masses, nodules, pleural effusion, or pneumothorax). The distinction between minor and major revisions hinged on whether the descriptions adequately covered clinically significant referable abnormalities. A report requiring minor revision might accurately list all referable abnormalities but need slight adjustments in shape, size, or location. Conversely, a report would require major revision if it mentioned several referable abnormalities but only partially described them. A report was considered unacceptable if it failed to mention any referable abnormalities, with potentially serious clinical implications. We defined successful autonomous reporting as reports rated either A) acceptable without any revision or B) acceptable with minor revisions.

**Statistical analysis**

The model's performance in generating radiologic reports was assessed using precision, recall, and F1 scores. The CheXpert-Labeler assigns "uncertain" labels to pathologic findings not mentioned in the report, and we excluded all uncertain labels from our analysis. We included only definite positive or negative labels. Additionally, due to the scarce number of images with labels such as "pleural other" and "fractures," these were omitted from the analysis. The specific criteria for removing certain labels and the details of the excluded labels are outlined in the accompanying table. To estimate the confidence intervals of the F1 scores, we utilized non-parametric bootstrapping with 1,000 iterations. For the evaluation conducted by human radiologists, the Cochran Q test was employed to determine the statistical significance of differences between the evaluations made by human radiologists and the model.

# Results

**Model performance on the internal test set**

Table 2 illustrates the report generation capabilities of our model on the MIMIC internal test set. The model achieved an average F1 score of 0.81 for six pathological labels, including cardiomegaly, consolidation, edema, pleural effusion, pneumonia, and pneumothorax. It demonstrated strong performance, with F1 scores exceeding 0.8, in identifying cardiomegaly, edema, and pleural effusion, while its ability to detect pneumothorax was weaker. Overall, the model exhibited higher average F1 scores than GPT-4-vision or Gemini-Pro-Vision.

Table 3 presents the model's pathology detection performance on the CheXpert internal test set. The model achieved an average F1 score of 0.57 for five pathological findings: atelectasis, cardiomegaly, consolidation, edema, and pleural effusion. While it performed relatively well in identifying lung opacity, atelectasis, and pleural effusion, its effectiveness in detecting consolidation was lower. This average F1 score of 0.57 is marginally lower than that of CheXzero, which achieved 0.61, and slightly below the 0.62 F1 score reported for human radiologists. No established F1 scores from CheXzero and human radiologists are available for diagnosing lung opacity and support devices, but our model demonstrated commendable F1 scores in detecting these conditions in CXR.

Figure 3 displays an example CXR, highlighting the format of the generated radiologic report. This report effectively pinpoints critical findings, yet it occasionally overlooks specific details, such as the presence of a central catheter.

**Model performance on the external test set**

In the external test set, the model produced an average F1 score of 0.62 for detecting cardiomegaly, consolidation, edema, lung opacity, pleural effusion, pneumonia, and pneumothorax. It showed an excellent ability to detect cardiomegaly, edema, and lung opacity, but its performance in detecting pneumothorax was significantly weaker (Table 4). Overall, the model outperformed other models in this regard. A review of several examples showed that the model accurately identified and described the corresponding lesions (Figures 4 and 5).

In the evaluation of radiologic report acceptability by human radiologists, the model achieved an "acceptable without any revision" rate of 51.3%, which closely aligns with the 54.0% acceptability rate of ground truth reports. To gauge the model's capability for autonomous reporting without human radiologist intervention, we defined successful autonomous reporting as reports deemed acceptable either without any revision or with only minor revisions. By this criterion, the model reached a success rate of 72.7%. While this is lower than the 84.0% success rate of ground truth reports, the difference in the rate of autonomous reporting between the model and the ground truth was found to be statistically significant, indicating that the model was somewhat inferior in terms of the autonomous reporting rate. However, the model still maintained a commendable success rate of over 70%.

# Discussion

We successfully developed a multimodal large language model capable of accurately detecting major pathological findings in CXRs and generating free-text radiologic reports. Our model exhibited relatively good performance compared to other publicly available general-purpose multimodal LLMs, such as GPT-4-vision and Gemini-Pro-Vision. We also explored the potential of multimodal LLMs for autonomous or semi-autonomous reporting in chest radiography. However, there are some limitations to our study.

First, the evaluation method we employed has inherent limitations. While we used CheXpert-Labeler to assess the quality of the reports, this tool only evaluates the presence of pathological labels and does not consider the location or number of pathological lesions. As a result, this method may not fully reflect the true accuracy of the generated reports. Second, our model showed poor performance in identifying certain pathological lesions, such as pneumothorax and consolidation. Notably, its diagnostic performance was inferior to that of human radiologists, as shown in the CheXpert internal test set. This might be partly due to the resolution limitations of our model, which processes 512×512 pixel images, a lower resolution than the higher-resolution images used by radiologists on specialized monitors. Moreover, our model processes 8-bit images with a grayscale of 256 levels, whereas radiologist monitors can display up to 10 or 12-bit grayscale images, providing finer details. These factors could contribute to the model's suboptimal performance in detecting subtle lesions. Third, the models we compared ours with are general-purpose and not fine-tuned for CXR interpretation. Therefore, it is not unexpected that our fine-tuned model would outperform them. Nevertheless, it is noteworthy that these general-purpose models still achieved high F1 scores in diagnosing conditions like cardiomegaly and lung opacity. Several non-peer-reviewed public multimodal LLMs, such as Xraygpt, UniXGen, and LLM-CXR, have been released for CXR interpretation, but we did not include them in our comparison due to potential dataset overlap, as we utilized a public dataset for both training and testing. Lastly, our assessment of the potential of our model for autonomous reporting was based on a limited dataset of just 50 CXRs, which does not mirror real-world clinical settings. Future research should involve larger-scale studies to ensure the safety and efficacy of multimodal LLMs in CXR interpretation.

In conclusion, our study demonstrates the capability of multimodal LLMs to generate radiologic reports that accurately recognize major lesions. By making our model open-source, we aim to promote the development of more capable and accurate models. We are confident that multimodal large language models have considerable potential to assist clinicians, reduce the workload of radiologists in clinical settings, and ultimately improve patient outcomes.

## Data Availability

This study utilized publicly available datasets. The links for downloading these datasets are provided in the table below.

| Dataset | URL |
| --- | --- |
| **BrixIA COVID-19 dataset** [21] | https://brixia.github.io/ |
| **CheXpert train/validation dataset** [15] | https://stanfordmlgroup.github.io/competitions/chexpert/ |
| **NIH dataset** [22] | https://nihcc.app.box.com/v/ChestXray-NIHCC |
| **PadChest dataset** [23] | http://bimcv.cipf.es/bimcv-projects/padchest/ |
| **RSNA COVID-19 AI Detection Challenge** [24] | https://www.rsna.org/rsnai/ai-image-challenge/covid-19-ai-detection-challenge-2021 |
| **VinDR dataset** [25] | https://vindr.ai/datasets/cxr |
| **MIMIC dataset** [14] | https://physionet.org/content/mimiciii/1.4/ |
| **CheXpert test dataset** [15] | https://stanfordaimi.azurewebsites.net/datasets/23c56a0d-15de-405b-87c8-99c30138950c |
| **CheXpert train/validation dataset** [15] | https://stanfordmlgroup.github.io/competitions/chexpert/ |
| **Indiana University dataset** [16] | https://openi.nlm.nih.gov/faq#collection |

## Code Availability

The structure and weights of the model used in this study are openly accessible to facilitate further research and development in the field. The code is available at https://github.com/ECOFRI/CXR_LLAVA.

# Tables

Table 1. Countries of collection, years of publication, and numbers of frontal chest radiographs in the publicly available datasets used for model training and evaluation.

| Dataset | Country of Collection | Year of Publication | Numbers of Frontal CXRs | | |
| --- | --- | --- | --- | --- | --- |
| | | | Training | Validation | Test |
| **Training dataset 1: chest radiograph datasets with pathologic findings labeled** | | | | | |
| BrixIA COVID-19 dataset [21] | Italy | 2021 | 3,755 | 470 | - |
| CheXpert train/validation dataset [15] | USA | 2019 | 152,983 | 19,123 | - |
| NIH dataset [22] | USA | 2017 | 70,671 | 8,833 | - |
| PadChest dataset [23] | Spain | 2019 | 86,438 | 10,805 | - |
| RSNA COVID-19 AI Detection Challenge [24] | Various countries | 2021 | 5,066 | 634 | - |
| VinDR dataset [25] | Vietnam | 2020 | 14,314 | 1,789 | - |
| *Subtotal* | | | 333,227 | 41,654 | - |
| **Training dataset 2: chest radiograph dataset with free-text radiologic reports** | | | | | |
| MIMIC dataset [14] | USA | 2019 | 193,513 | 24,186 | - |
| **Internal test sets** | | | | | |
| MIMIC dataset (randomly selected) [14] | USA | 2019 | - | - | 3,000 |
| CheXpert test dataset [15] | USA | 2022 | - | - | 518 |
| *Subtotal* | | | - | - | 3,518 |
| **External test set** | | | | | |
| Indiana University dataset [16] | USA | 2016 | - | - | 3,689 |

**Table 2. Model performance with the MIMIC internal test set.**

The model achieved an excellent average F1 score of 0.81, outperforming the GPT-4-vision and Gemini-Pro-Vision models, which scored 0.62 and 0.68, respectively.

| F1 scores of each pathologic label in the MIMIC internal test set | | | |
|---|---|---|---|
| **Models** | **CXR-LLAVA** | **GPT-4-vision** | **Gemini-Pro-Vision** |
| **Cardiomegaly** | 0.86 (0.85, 0.88) | 0.77 (0.75, 0.79) | 0.78 (0.76, 0.80) |
| **Consolidation** | 0.68 (0.57, 0.78) | 0.20 (0.11, 0.29) | 0.41 (0.36, 0.47) |
| **Edema** | 0.84 (0.81, 0.87) | 0.71 (0.63, 0.78) | 0.69 (0.66, 0.72) |
| **Pleural effusion** | 0.83 (0.81, 0.85) | 0.39 (0.35, 0.43) | 0.61 (0.58, 0.63) |
| **Pneumonia** | 0.65 (0.54, 0.74) | 0.79 (0.73, 0.84) | 0.82 (0.77, 0.86) |
| **Pneumothorax** | 0.46 (0.37, 0.53) | 0.03 (0.00, 0.07) | 0.00 (0.00, 0.00) |
| **Average** | 0.81 (0.80, 0.82) | 0.62 (0.61, 0.64) | 0.68 (0.66, 0.69) |

Note: In our analysis, specific labels such as "lung lesion," "lung opacity," "atelectasis," "pleural other," "fracture," and "support devices" were excluded due to their low frequency, being under 5% in either the negative or positive class or having a sample number below 10, which makes them statistically less significant for a balanced analysis. Additionally, the label "enlarged cardiomediastinum" was not included as it significantly overlaps with "cardiomegaly," which could lead to redundant data interpretations.

**Table 3. Model performance with the CheXpert internal test set.**

The model attained an average F1 score of 0.57 for five key pathologies, which was marginally lower than CheXzero (0.61) and human radiologists (0.62). However, it demonstrated exceptional capability in identifying lung opacity, support devices, and atelectasis.

| F1 scores of each pathologic label in the CheXpert internal test set | | | | | |
|---|---|---|---|---|---|
| **Models** | **CXR-LLAVA** | **GPT-4-vision** | **Gemini-Pro-Vision** | **CheXzero[20]** | **Human Radiologists[20]** |
| **Atelectasis** | 0.69 (0.64, 0.74) | 0.00 (0.00, 0.00) | 0.04 (0.00, 0.08) | 0.65 (0.59, 0.70) | 0.69 (0.65, 0.73) |
| **Cardiomegaly** | 0.62 (0.56, 0.67) | 0.46 (0.42, 0.51) | 0.39 (0.33, 0.45) | 0.74 (0.69, 0.79) | 0.68 (0.63, 0.72) |
| **Consolidation** | 0.24 (0.17, 0.31) | 0.00 (0.00, 0.00) | 0.11 (0.07, 0.15) | 0.33 (0.24, 0.42) | 0.39 (0.28, 0.49) |
| **Edema** | 0.50 (0.43, 0.57) | 0.05 (0.00, 0.11) | 0.19 (0.11, 0.26) | 0.60 (0.52, 0.68) | 0.58 (0.51, 0.65) |
| **Pleural effusion** | 0.63 (0.57, 0.69) | 0.41 (0.34, 0.47) | 0.03 (0.00, 0.09) | 0.70 (0.63, 0.76) | 0.74 (0.69, 0.78) |
| **Average for the 5 pathologies** | 0.57 (0.54, 0.59) | 0.35 (0.32, 0.39) | 0.19 (0.17, 0.22) | 0.61 (0.57, 0.64) | 0.62 (0.59, 0.64) |
| **Lung opacity** | 0.84 (0.81, 0.87) | 0.71 (0.66, 0.75) | 0.68 (0.64, 0.72) | N/A | N/A |
| **Support devices** | 0.78 (0.74, 0.82) | 0.72 (0.69, 0.76) | 0.70 (0.66, 0.74) | N/A | N/A |
| **Overall average** | 0.67 (0.65, 0.69) | 0.53 (0.51, 0.55) | 0.45 (0.43, 0.47) | N/A | N/A |

Note: unbalanced labels like "lung lesion," "pneumonia," "pneumothorax," "pleural other," and "fracture," which had a frequency of less than 5% in either the negative or positive class, were not included in the analysis. Additionally, the label "enlarged cardiomediastinum" was not included as it significantly overlaps with "cardiomegaly," which could lead to redundant data interpretations.

**Table 4. Model performance with the Indiana external test set.**

The model achieved an overall average F1 score of 0.62, excelling particularly in the detection of cardiomegaly (0.62), edema (0.67), and lung opacity (0.85). However, its performance in detecting pneumothorax was notably lower (0.05).

| F1 scores of each pathologic label in the Indiana external test set | | | |
|---|---|---|---|
| **Models** | **CXR-LLAVA** | **GPT-4-vision** | **Gemini-Pro-Vision** |
| **Cardiomegaly** | 0.62 (0.57, 0.65) | 0.37 (0.34, 0.39) | 0.39 (0.37, 0.42) |
| **Consolidation** | 0.31 (0.09, 0.50) | 0.08 (0.00, 0.17) | 0.07 (0.03, 0.11) |
| **Edema** | 0.67 (0.33, 0.86) | 0.25 (0.00, 0.52) | 0.28 (0.18, 0.37) |
| **Lung Opacity** | 0.85 (0.80, 0.89) | 0.58 (0.54, 0.62) | 0.00 (0.00, 0.00) |
| **Pleural Effusion** | 0.55 (0.48, 0.62) | 0.13 (0.08, 0.18) | 0.17 (0.14, 0.20) |
| **Pneumonia** | 0.63 (0.42, 0.79) | 0.63 (0.45, 0.77) | 0.82 (0.56, 0.96) |
| **Pneumothorax** | 0.05 (0.00, 0.13) | 0.00 (0.00, 0.00) | 0.00 (0.00, 0.00) |
| **Average** | 0.62 (0.59, 0.65) | 0.39 (0.37, 0.41) | 0.30 (0.29, 0.32) |

Note: we excluded labels that were unbalanced, with fewer than 10 samples in either the negative or positive category. This included labels such as "lung lesion," "atelectasis," "pleural other," "fracture," and "support devices." These were omitted to ensure statistical relevance and balance in the analysis. Additionally, the label "enlarged cardiomediastinum" was not included as it significantly overlaps with "cardiomegaly," which could lead to redundant data interpretations.

**Table 5. Evaluation of radiologic report acceptability by human radiologists from the Indiana external test set.**

The model achieved a 51.3% rate (77 cases) of being "acceptable without any revision" (Class A), closely mirroring the 54.0% rate of the ground truth reports. The model's success rate for autonomous reporting (Class A+B) reached 72.7% (109 cases), slightly lower than the 84.0% for ground truth reports. This difference was statistically significant ($p<0.001$), highlighting the comparative capabilities and limitations of the model in autonomous radiologic reporting.

| Class | Meaning | CXR-LLAVA | Ground Truth | Comparison |
|---|---|---|---|---|
| A | Acceptable without any revision | 77 (51.3%) | 81 (54.0%) | |
| B | Acceptable after minor revision | 32 (21.3%) | 45 (30.0%) | |
| C | Acceptable after major revision | 8 (5.3%) | 6 (4.0%) | |
| D | Unacceptable | 33 (22.0%) | 18 (12.0%) | |
| A+B | Successful autonomous reporting | 109 (72.7%) | 126 (84.0%) | p<0.001 |

# Figures

**Figure 1. CXR-LLAVA training process.** (A) Initially, the image encoder was trained on a basic classification task to differentiate between normal and abnormal CXRs, thereby acquiring fundamental representations of CXRs. (B) Subsequently, the model underwent training with pairs of CXRs and their corresponding pathological findings. This training employed the contrastive language-image pre-training (CLIP) strategy to foster shared representations between images and text. (C) The image encoder was then assimilated into CXR-LLAVA, initiating the alignment of image representations with the large language model (LLM). In this phase, training focused on pairs of CXR images and radiologic reports, with updates confined to the projection layer. (D) Upon successful alignment of the image encoder with the LLM, an instruction fine-tuning process was undertaken. This involved a variety of radiologic reports and question-answer pairs, aiming to refine the model's capability to interpret CXRs and facilitate more informative interactions. Please note that the figure abstracts from the detailed neural network information, omitting elements such as tokenizer, batch normalization, projection, and linear classification layers.

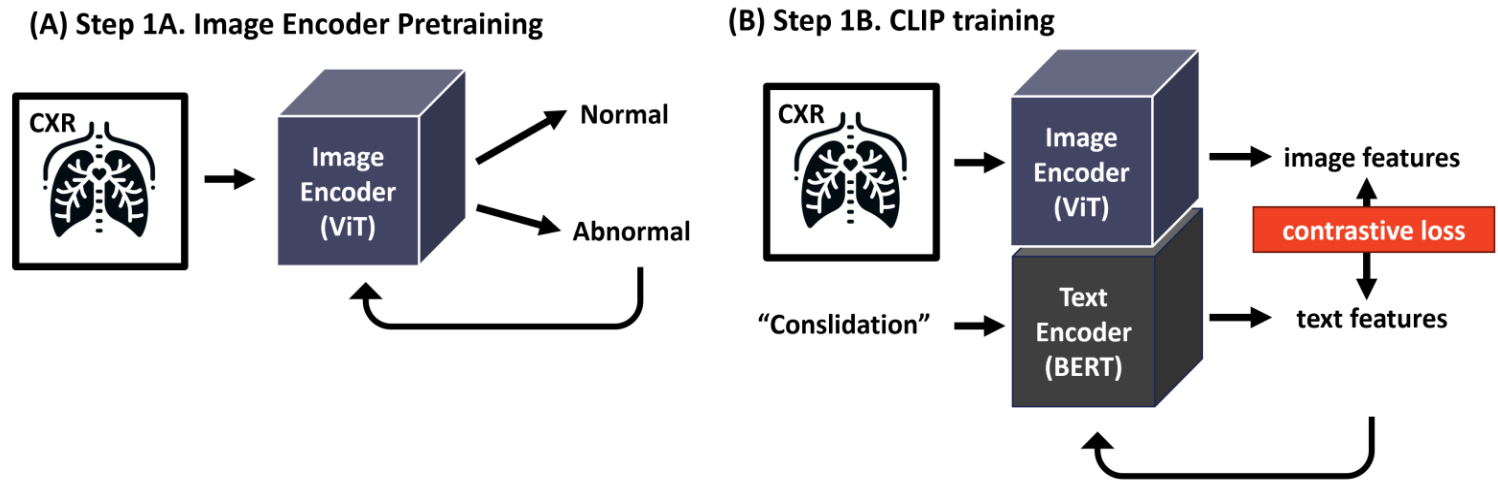

**(C) Step 2A. Image encoder feature alignment**

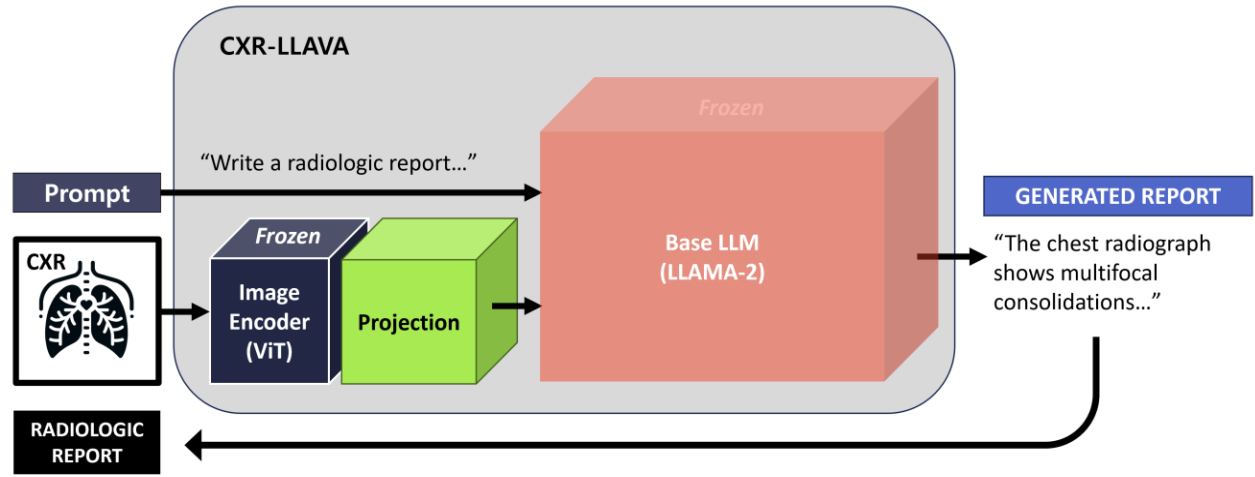

**(D) Step 2B. Multimodal LLM instruction fine tuning**

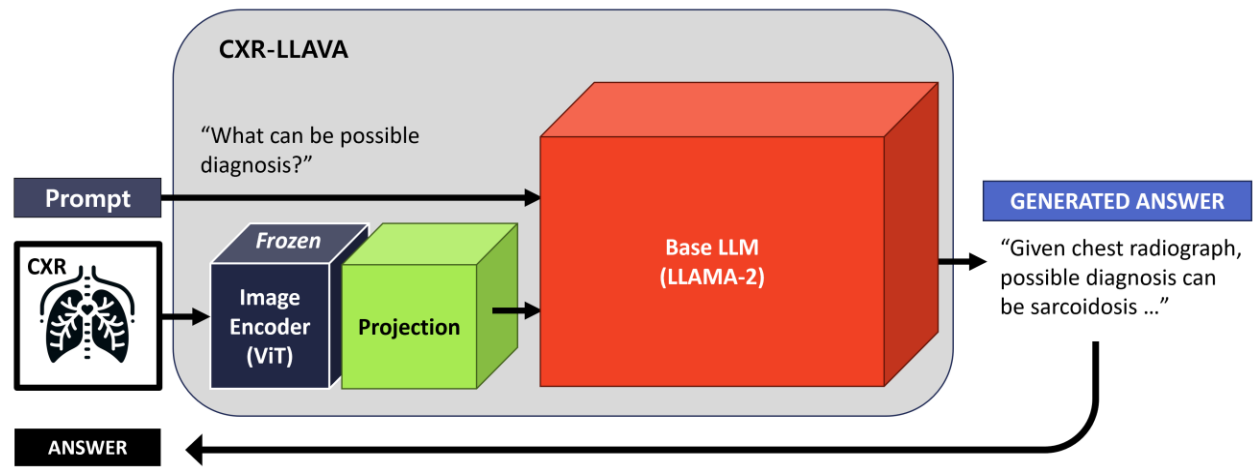

**Figure 2. Model evaluation flow diagram**

(a) Evaluation of datasets with ground-truth free-text radiologic reports, specifically the MIMIC internal test set and the Indiana external test set. Pathologic labels were obtained using the CheXpert labeler from both the original reports and the model-generated reports, with a subsequent comparison of these results.

(b) Evaluation of datasets with established ground-truth pathologic labels, specifically the CheXpert internal test set, involved directly generating pathologic labels from the model using a label generation prompt.

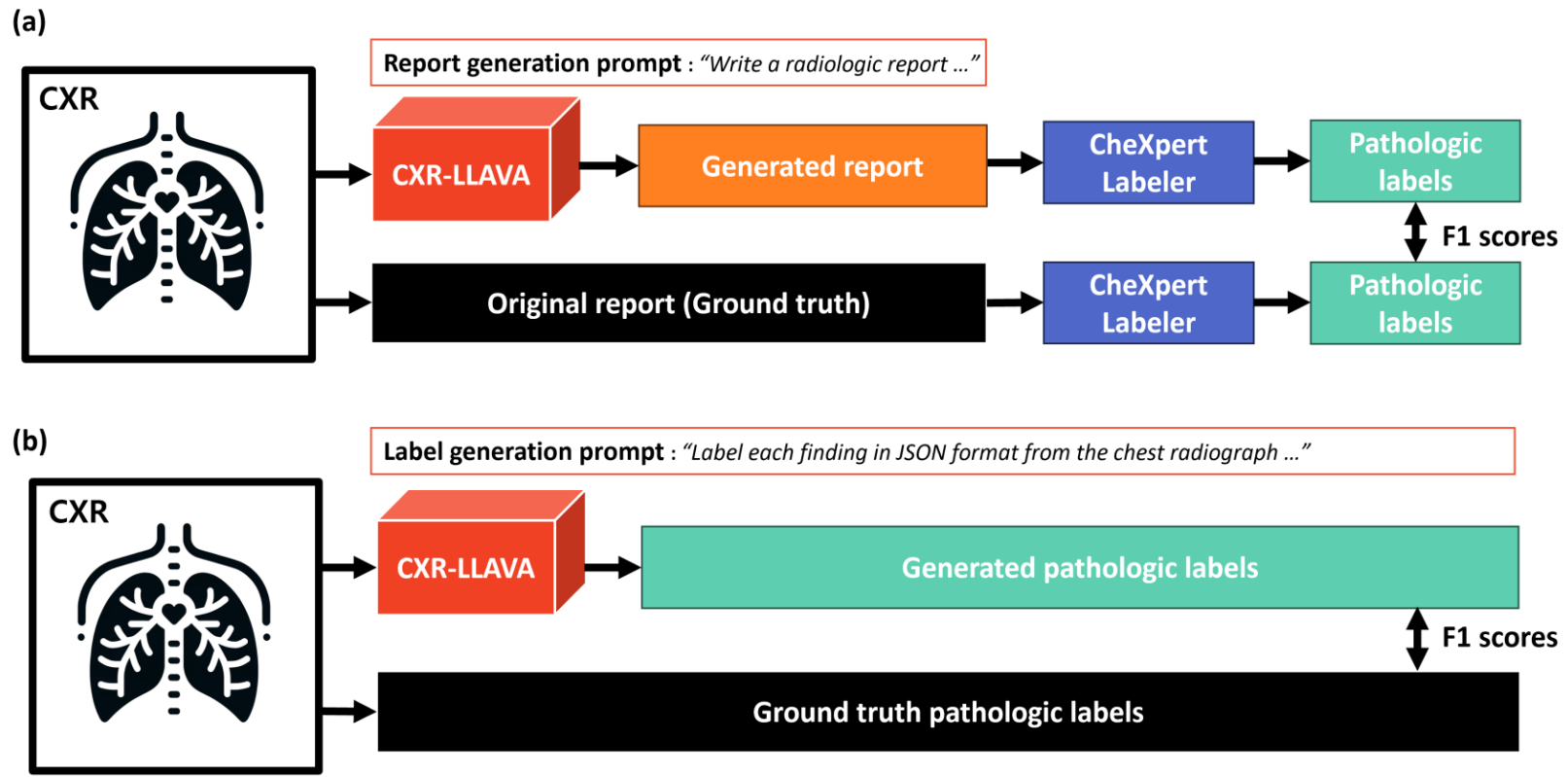

**Figure 3. An example of a chest radiograph from the CheXpert internal test set.**

While the model identified the presence of pleural effusions, atelectasis, and lung opacity, it omitted details about the central catheter (support device).

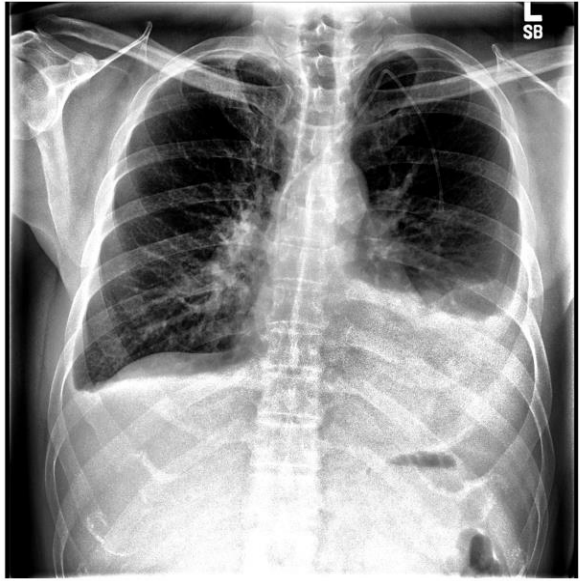

**Figure 4. An example of a chest radiograph from the Indiana external test set.**

The model's interpretation included information about bilateral pulmonary nodules and suggested a possible diagnosis of lung metastasis or infection, which is reasonable.

It also recommended that an additional chest CT scan might be helpful. However, the model could not detect the implanted venous access device.

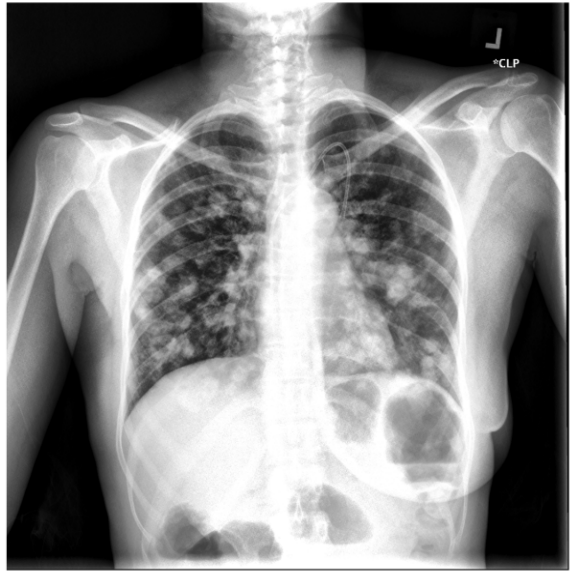

[Question1] Write a radiologic report on the given chest radiograph, including information about atelectasis, cardiomegaly, consolidation, pulmonary edema, pleural effusion, and pneumothorax.

[CXR-LLAVA] The chest radiograph reveals numerous bilateral pulmonary nodules. There is no evidence of pneumothorax. The heart size appears normal.

[Question2] What are the possible differential diagnoses for this patient?

[CXR-LLAVA] Possible differential diagnoses for this patient include metastatic prostate cancer, lymphoma, and infectious diseases such as tuberculosis or fungal infections.

[Question3] What kind of further examination is required?

[CXR-LLAVA] A chest CT scan is recommended for a more detailed assessment of the pulmonary nodules and potential metastases.

[Ground truth]

Findings: The heart is normal in size. The mediastinum is stable. Left-sided chest XXXX is again visualized with tip at cavoatrial junction. There is no pneumothorax. Numerous bilateral pulmonary nodules have increased in size and number XXXX compared to prior study. The dominant nodule/mass in the left midlung is also mildly increased. There is no pleural effusion.

Impression : Interval increase in size and number of innumerable bilateral pulmonary nodules consistent with worsening metastatic disease.

**Figure 5. An example of a chest radiograph from the Indiana external test set.**

The model's interpretation identified right upper lobe consolidation and proposed pneumonia as a possible diagnosis, which is reasonable. Nonetheless, the model failed to detect a small left upper lung nodule (black arrow).

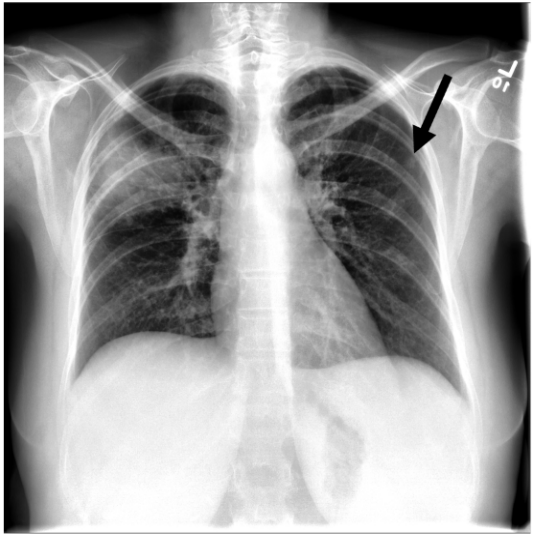

[Question1] Write a radiologic report on the given chest radiograph, including information about atelectasis, cardiomegaly, consolidation, pulmonary edema, pleural effusion, and pneumothorax.

[CXR-LLAVA] The chest radiograph reveals a consolidation in the right upper lobe of the lungs. There is no evidence of pneumothorax or pleural effusion. The cardiac contour appears normal.

[Question2] What can be possible diagnosis?

[CXR-LLAVA] Based on the imaging, the consolidation could be due to pneumonia. However, further diagnostic tests such as a sputum culture or blood tests may be needed to confirm the diagnosis.

[Question3] What kind of further examination is required?

[CXR-LLAVA] A CT scan of the chest is recommended for a more detailed evaluation of the possible diagnoses.

[Ground truth]

Findings: The cardiomediastinal silhouette and vasculature are within normal limits for size and contour. There is right upper lobe airspace disease. There is a rounded nodular opacity in the left upper lung measuring approximately 7 mm which may represent further sequela of infectious process versus other pathology. Osseous structures are within normal limits for patient age.

Impression: 1. Right upper lobe pneumonia. 2. Rounded nodular opacity in the peripheral left upper lung which may represent further sequela infectious process versus other pathology including metastatic disease in a patient with thyroid cancer. Follow up to resolution recommended.

# Acknowledgements

We appreciate the high-performance GPU computing support of HPC-AI Open Infrastructure via GIST SCENT. We also acknowledge the utilization of LLM (gpt-4, OpenAI) to enhance the quality of our medical writing. Any revisions made by LLM was thoroughly reviewed by the authors and subsequent adjustment were made as deemed appropriate.

This work was partly supported by Institute of Information and Communications Technology Planning and Evaluation (IITP) grant funded by the Korea government (MSIT) [No.2019-0-01842, Artificial Intelligence Graduate School Program (GIST), No. 2021-0-02068, Artificial Intelligence Innovation Hub], the National Research Foundation of Korea under grant NRF-2022R1F1A1068529.

# Supplementary Materials

## Detailed Architecture and Training Process of the CXR Image Encoder

We have developed a CXR image encoder utilizing the vision transformer architecture [1]. Empirically, "ViT-L/16" was selected to balance computational cost and performance. Following the training process, this image encoder transforms CXR images (i.e., those with dimensions of 512x512 and single-channel) into a 128-dimensional representation vector, which is subsequently employed for further analysis. Our approach employs a two-staged training strategy to impart radiological context.

In the initial training phase, we first pretrained the image encoder to distinguish between normal and abnormal images. This was accomplished by adding a dense binary classifier to the image encoder. Specifically, we used Dataset 1, and the implementation details are as follows. The model was initialized with random weights. It was trained for up to 100 epochs with a learning rate of 1e-3 using the SGD optimizer and a batch size of 64. Training was conducted for approximately two weeks on a single GPU (NVIDIA A100). The final model was chosen based on the lowest validation loss, and the trained model showed an area under curve of receptor operator characteristics of approximately 0.92 for the binary classification task.

Subsequent to the initial training phase, we incorporated the CLIP (Contrastive Language-Image Pretraining) method to further enhance the image encoder's ability to understand the relationship between text and CXR images [2]. Specifically, we utilized a text encoder based on the Bidirectional Encoder Representations from Transformers (BERT) from prior research [3]. The trained image encoder was then integrated and trained to minimize the contrastive loss between the image vector and text vector. In this step, we trained the encoders twice with different datasets (Dataset 1 and 2). For the initial training stage with Dataset 1, we used pathology labels that lacked location information for learning representation. To accelerate the learning speed, the text encoder was frozen up to step 70k, and then it was unfrozen, and training continued. A learning rate of 1e-2 with an SGD optimizer was used during this phase, and it took about a day using eight NVIDIA A100s. Training was halted when the validation loss reached a plateau. In the subsequent training stage with Dataset 2, radiologic reports, containing not only pathology but also the locations of the pathology, were used to learn the relationship between text and CXR images. The model that demonstrated the lowest validation loss was ultimately selected.

## Detailed Architecture and Training Process of the CXR-LLaVA.

We adopted the concept of LLaVA [4] for CXR-LLaVA, which comprises an image encoder trained in the preceding process, a multimodal projection layer, and LLAMA2-7B-CHAT. When a text prompt and a CXR image are input into this model, the LLAMA2 tokenizer initially converts the text prompt into a vector. Simultaneously, the image encoder processes the CXR, producing 128x1024-dimensional image tokens from 512x512 images. These tokens are then transformed into a 5120-dimensional vector through the dense multimodal projection layer. The resulting 5120-dimensional vector is inserted into the specified location within the text prompt. Once it's fed autoregressively into the causal LLM, a response is generated.

The training process for CXR-LLaVA occurred in two stages. In the first stage, all layers except the multimodal projection layer were frozen. Training was then conducted to align the image and text vector spaces. This phase lasted for 1 epoch, using a learning rate of 2e-3 with the Adam optimizer, a batch size of 16, and took about 4 hours on eight NVIDIA A100 40GB GPUs. In the second stage, only the image encoder was frozen, while all other layers were set to be trainable, continuing the training. This stage was conducted over 3 epochs with a learning rate of 2e-5, using the Adam optimizer, a batch size of 16, and took approximately 50 hours on the same GPU setup. No validation was performed during the CXR-LLaVA training process, and the final model iteration was selected as the ultimate model.

# Refining the Radiologic Report and Preparing the LLM Fine-Tuning Dataset

The MIMIC dataset provides free-text radiology reports for CXRs. However, original radiology reports are not suitable for use as LLAVA training data since they contain contextual information that cannot be inferred from a single CXR, like comparisons with previous images and the patient's medical history. We used OpenAI GPT-4 to remove such parts, and the prompt we used is as follows:

---

You are skillful radiologist and doing summarization of chest x-ray report.
Summarize these information from the report.
Answer to each questions as json format which have "standard report", "conclusion" and "recommendation" as keys.

1. "standard_report" : Write a standardized radiologic report as one paragraph. Standardized report must include information about abnormality of lungs, mediastinum, heart and thorax.
2. "conclusion" : What is the conclusion or impression of the radiologic report? Include only critical information.
3. "recommendation" : Should additional radiologic study needed? What type of study should be performed?

Do not include any temporal or time information in standard_report and conclusion. DO NOT USE WORD SUCH AS "new", "previous", "comparison", "stable", "improved", "improving", "decreased", "increased", "changed", "unchanged", "resolved", or "cleared".
Do not include information about 'comparison with prior study'.
Do not include information about lateral radiograph.
Replace any numeric information, such as millimeter or centimeter
Remove any information about patient age, gender, and medical history.
Remove any under-bar & blank.
Remove any information or location about catheter, chest tube, endotracheal tube, PICC, chemoport, central line, nasogastric tube or other medical devices.

---

In addition to simply generating radiologic report, we also created a Q&A dataset for each chest radiograph to incorporate a question answering feature as follows:

---

"question1" : Compose a question from the perspective of a student radiologist, inquiring about the anatomical location, number, or presence of pathology in the chest radiograph.
"answer1" : Write an informative answer to question1.
"question2" : Compose a question that asks possible differential diagnoses from this chest radiograph, without referring to the patient's history.
"answer2" : Write an informative answer to question2.

# Label Distribution in the MIMIC Internal Test Set

The MIMIC internal test set provides labels for 14 findings. Their distribution is as follows:

|  | Negative | Positive |
|---|---|---|
| **Enlarged Cardiomediastinum** | 661 | 114 |
| **Cardiomegaly** | 255 | 756 |
| **Lung Opacity** | 2 | 17 |
| **Lung Lesion** | 11 | 369 |
| **Edema** | 416 | 53 |
| **Consolidation** | 175 | 255 |
| **Pneumonia** | 1183 | 703 |
| **Atelectasis** | 12 | 416 |
| **Pneumothorax** | 63 | 73 |
| **Pleural Effusion** | 1206 | 173 |
| **Pleural Other** | 0 | 3 |
| **Fracture** | 0 | 7 |
| **Support Devices** | 3 | 134 |

## Label Distribution in the CheXpert Internal Test Set

The CheXpert internal test set provides labels for 14 findings. Their distribution is as follows:

|  | Negative | Positive |
|---|---|---|
| **No Finding** | 450 | 68 |
| **Enlarged Cardiomediastinum** | 262 | 256 |
| **Cardiomegaly** | 364 | 154 |
| **Lung Opacity** | 246 | 272 |
| **Lung Lesion** | 509 | 9 |
| **Edema** | 439 | 79 |
| **Consolidation** | 489 | 29 |
| **Pneumonia** | 507 | 11 |
| **Atelectasis** | 360 | 158 |
| **Pneumothorax** | 509 | 9 |
| **Pleural Effusion** | 413 | 105 |
| **Pleural Other** | 518 | 0 |
| **Fracture** | 513 | 5 |
| **Support Devices** | 252 | 266 |

# Label Distribution in the Indiana External Test Set

The Indiana external test set provides free-text radiology reports. We extracted labels from radiology reports using CheXpert-Labeler. Their distribution is as follows:

|  | **Negative** | **Positive** |
|---|---|---|
| **Enlarged Cardiomediastinum** | 1105 | 84 |
| **Cardiomegaly** | 727 | 371 |
| **Lung Lesion** | 3 | 7 |
| **Consolidation** | 285 | 15 |
| **Edema** | 109 | 14 |
| **Lung Opacity** | 68 | 154 |
| **Pleural Effusion** | 2295 | 118 |
| **Atelectasis** | 1 | 77 |
| **Pneumonia** | 57 | 23 |
| **Pneumothorax** | 1427 | 27 |
| **Pleural Other** | 0 | 1 |
| **Fracture** | 1 | 1 |
| **Support Devices** | 3 | 9 |